\documentclass[10pt,twocolumn,letterpaper]{article}

\usepackage{iccv}
\usepackage{times}
\usepackage{epsfig}
\usepackage{graphicx}
\usepackage{amsmath}
\usepackage{amssymb}

\newcommand{\PAR}[1]{\vskip4pt \noindent{\bf #1~}}

\newcommand{\wb}[1]{{\color{black}{#1}}}

\usepackage{makecell}
\usepackage{dsfont}
\usepackage{enumitem}
\usepackage{subfigure}
\usepackage{caption}
\usepackage{arydshln}
\usepackage{comment}
\usepackage{color}
\usepackage[dvipsnames, svgnames, x11names]{xcolor}
\usepackage{amsfonts}
\usepackage{url}
\usepackage{gensymb}
\usepackage{bm}
\usepackage{enumitem}
\usepackage{textcomp}
\usepackage[normalem]{ulem}
\useunder{\uline}{\ul}{}
\usepackage{multirow}
\usepackage{multicol}
\DeclareMathOperator*{\argmax}{arg\,max}

\usepackage[breaklinks=true,bookmarks=false]{hyperref}

\iccvfinalcopy % *** Uncomment this line for the final submission
\def\iccvPaperID{4191} % *** Enter the ICCV Paper ID here

\ificcvfinal\fi

\begin{document}

%%%%%%%%% TITLE
\title{P2-Net: Joint Description and Detection of Local Features for \\
       Pixel and Point Matching}
\author{Bing Wang$^1$, Changhao Chen$^2$, Zhaopeng Cui$^3$, Jie Qin$^4$, Chris Xiaoxuan Lu$^5$,\\ 
Zhengdi Yu$^6$, Peijun Zhao$^1$, Zhen Dong$^7$, Fan Zhu$^4$, Niki Trigoni$^1$, Andrew Markham$^1$\\
\vspace{-0.2cm}
\\
\normalfont
\normalsize\textsuperscript{$1$}University of Oxford\quad
\normalsize\textsuperscript{$2$}National University of Defense Technology\quad
\normalsize\textsuperscript{$3$}Zhejiang University\quad 
\quad
\\
\normalfont
\normalsize\textsuperscript{$4$}Inception Institute of Artificial Intelligence\quad
\normalsize\textsuperscript{$5$}University of Edinburgh\quad
\normalsize\textsuperscript{$6$}Durham University\quad
\normalsize\textsuperscript{$7$}Wuhan University
}
\maketitle
\ificcvfinal\thispagestyle{empty}\fi

%%%%%%%%% ABSTRACT
\begin{abstract}
Accurately describing and detecting 2D and 3D keypoints is crucial to establishing correspondences across images and point clouds. Despite a plethora of learning-based 2D or 3D local feature descriptors and detectors having been proposed, the derivation of a shared descriptor and joint keypoint detector that directly matches pixels and points remains under-explored by the community. This work takes the initiative to establish \emph{fine-grained} correspondences between 2D images and 3D point clouds. In order to directly match pixels and points, a dual fully-convolutional framework is presented that maps 2D and 3D inputs into a shared latent representation space to simultaneously describe and detect keypoints. Furthermore, \wb{an ultra-wide reception mechanism and a novel loss function are} designed to mitigate the intrinsic information variations between pixel and point local regions. Extensive experimental results demonstrate that our framework shows competitive performance in fine-grained matching between images and point clouds and achieves state-of-the-art results for the task of indoor visual localization. Our source code is available at \url{https://github.com/BingCS/P2-Net}.
\end{abstract}
\vspace{-0.5cm}

%%%%%%%%% BODY TEXT
\section{Introduction}
\vspace{-0.0cm}
\label{sec: intro}
%background: direct match v.s. 2D->3D match 3D
% challenges: 
% 
Establishing accurate pixel- and point- level matches across images and point clouds, respectively, is a fundamental computer vision task that is crucial for a multitude of applications, such as Simultaneous Localization And Mapping \cite{montemerlo2002fastslam}, Structure-from-Motion \cite{schonberger2016structure}, pose estimation \cite{nath2018object}, 3D reconstruction \cite{heinly2015reconstructing}, and visual localization \cite{sattler2016efficient}.
%For such applications, extracted descriptors and detected keypoints should be discriminative, repeatable and robust to accurate matching under various scenarios with difficult lighting, repetition, motion blur, etc.

A typical pipeline of most existing methods is to first recover the 3D structure given an image sequence \cite{hartley2003multiple, salas2013slam++}, %(e.g. with photogrammetric techniques \cite{salas2013slam++}), 
and subsequently perform matching between pixels and points based on the 2D to 3D reprojected features. These features will be homogeneous as the points in the reconstructed 3D model inherit the descriptors from the corresponding pixels of the image sequence. However, \wb{this two-step procedure requires accurate 3D reconstruction, which is not always feasible to be achieved, e.g., under challenging illumination or large viewpoint changes.} More critically, this approach treats RGB images as ``first-class citizens'', and discounts the equivalence of sensors capable of directly capturing 3D point clouds, e.g., LIDAR, imaging RADAR and depth cameras. These factors motivate us to consider a unified approach to \textit{pixel and point matching}, where an open question can be posed: how to directly establish correspondences between pixels in 2D images and points in 3D point clouds, and vice-versa? This is inherently challenging as 2D images capture scene appearance, whereas 3D point clouds encode structure.

\begin{figure}
    \centering
    \includegraphics[width=0.93\columnwidth]{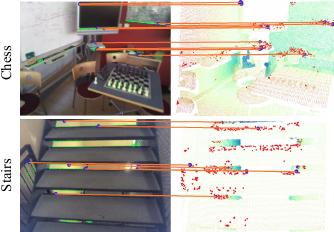}
    			    \vspace{-0.3cm}
    \caption{\wb{\textbf{Examples of 2D-3D matches obtained by the P2-Net.} The proposed method can directly establish correspondences across images and point clouds by the jointly learned feature description and detection.}}
    \label{fig:visualization}
\vspace{-0.5cm}
\end{figure}

%This problem is undoubtedly challenging for most
\wb{To this end, we formulate a new task of \textit{direct} 2D pixel and 3D point matching (\cf Fig.~\ref{fig:visualization}) \emph{without} any auxiliary steps (e.g., reconstruction).} This task is undoubtedly challenging for existing conventional and learning-based approaches, which fail to bridge the gap between 2D and 3D representations as separately extracted 2D and 3D local features are distinct and do not share a common embedding.
%, i.e., descriptors from images cannot be directly used in the 3D space and vice versa. 
Some recent works \cite{feng20192d3d,pham2020lcd} have attempted to associate descriptors from different domains by mapping 2D and 3D inputs onto a shared latent space. However, they only construct patch-wise descriptors, leading to coarse-grained matching results only. 

Even if fine-grained and accurate descriptors can be successfully obtained, direct pixel and point correspondences are still very difficult to establish. \textbf{First}, 2D and 3D keypoints are extracted based on distinct strategies - what leads to a good match in 2D (e.g., flat, visually distinct area such as a poster), does not necessarily correspond to what makes a strong match in 3D (e.g., a poorly illuminated corner of the room). Additionally, because of the sparsity of point clouds, a point local feature can be mapped to (or from) many pixel features taken from pixels that are spatially close to the point, increasing the matching ambiguity. \textbf{Second}, due to the large discrepancy between 2D and 3D data property and inflexible optimization manner, existing descriptor loss formulations \cite{dusmanu2019d2, luo2020aslfeat, bai2020d3feat} for either 2D or 3D local feature description do not guarantee convergence in this new context. Moreover, their detector designs only focus on penalizing the confounding descriptors from a safe region, incurring sub-optimal matching results in practice.

\wb{To tackle all these challenges, we propose a dual fully-convolutional framework, named \textbf{P}ixel and \textbf{P}oint Network (\textbf{P2-Net}), which is able to simultaneously achieve feature description and detection between 2D and 3D views. Furthermore, an \textbf{ultra-wide reception} mechanism is equipped when extracting descriptors to tackle the intrinsic information variations between pixel and point local regions. To optimize the network, we then design \textbf{P2-Loss}, consisting of two components: 1) a \textbf{circle-guided descriptor loss} in combination with a full sampling strategy, allowing to robustly learn distinctive descriptors by optimizing positive and negative matches in a self-paced manner; 2) a \textbf{batch-hard detector loss}, which additionally seeks for the repeatability of detections by encouraging the difference between the positive and globally hardest negative matches. Overall, our contributions are as follows:
\vspace{-0.2cm}
\begin{enumerate}[itemsep=-0.02mm]
    \item[1.] We propose a joint learning framework with an ultra-wide reception mechanism for simultaneous 2D and 3D local features description and detection to achieve direct pixel and point matching.
    \item[2.] We design a novel loss, composed of a circle-guided descriptor loss and a batch-hard detector loss, to robustly learn distinctive descriptors whilst explicitly guiding accurate detections for both pixels and points.
    \item[3.] We conduct extensive experiments and ablation studies, demonstrating the practicability of the proposed framework and the generalization ability of the new loss, and providing the intuition behind our choices.
\end{enumerate}
\vspace{-0.2cm}
}
To the best of our knowledge, this is the first joint learning framework to handle 2D and 3D local features description and detection for direct pixel and point matching.

%Experimental results show that our descriptor gives robust and promising results in all the tasks even when it is not purposely tailored to. Moreover, by the evaluation on image matching and point cloud tasks, the strong generalization ability of our proposed loss is also confirmed under different domains.

% \zp{I think the structure of introduction can be further improved as follows:
% \begin{itemize}
%     \item At first, claiming pixel-to-point matching is important for many tasks, like SLAM, visual localization...
%     \item Then, we mentioned that most existing 3d scene models are built together with the images in order to associate 2D features with 3D points. There is an open question - how to establish pixel-to-point correspondences given images and such 3D model (The current second paragraph)
%     \item Describe the challenges of such problem and why existing methods fail.
%     \item Describe how we solve these challenges. 
\vspace{-0.2cm}
\section{Related Work}
\vspace{-0.1cm}
%This section reviews the learning approaches for 
%local feature description and detection in 2D \cite{balntas2017hpatches, schonberger2017comparative} and 3D \cite{guo2016comprehensive, tombari2013performance} domain.

\subsection{2D Local Features Description and Detection}
\vspace{-0.1cm}
Previous learning-based methods in 2D domain simply replaced the descriptor \cite{tian2017l2, tian2019sosnet, liu2019gift, ebel2019beyond, pautrat2020online} or detector \cite{savinov2017quad, zhang2018learning, barroso2019key} with a learnable alternative. Recently, approaches to joint description and detection of 2D local features have attracted increased attention. 
LIFT \cite{yi2016lift} is the first, fully learning-based architecture to achieve this by rebuilding the main processing steps of SIFT with neural networks. Inspired by LIFT, SuperPoint \cite{detone2018superpoint} additionally tackles keypoint detection as a supervised task with labelled synthetic data before 
description, followed by being extended to an unsupervised version \cite{christiansen2019unsuperpoint}. Differently, DELF \cite{noh2017large} and LF-Net \cite{ono2018lf} exploit an attention mechanism and an asymmetric gradient back-propagation scheme, respectively, to enable unsupervised learning. Unlike previous research that separately learns the descriptor and detector, D2-Net \cite{dusmanu2019d2} designs a joint optimization framework based on non-maximal-suppression. To further encourage keypoints to be reliable and repeatable, R2D2 \cite{revaud2019r2d2} proposes a listwise ranking loss based on  differentiable average precision. Meanwhile, deformable convolution is introduced in ASLFeat \cite{luo2020aslfeat} for the same purpose. 
%Without directly learning for feature detection and description, Reinforced Feature Points \cite{bhowmik2020reinforced} propose a general training scheme based on policy gradient which can be applied to most aforementioned architectures for high-level visual localization task.
\vspace{-0.10cm}
\subsection{3D Local Features Description and Detection}
\vspace{-0.10cm}
Most prior work in the 3D domain has focused on the learning of descriptors. Instead of directly processing 3D data, early attempts \cite{su2015multi, zhou2018learning} instead extract a representation from multi-view images for 3D keypoint description. In contrast, 3DMatch \cite{zeng20173dmatch} and PerfectMatch \cite{gojcic2019perfect} construct descriptors by converting 3D patches into a voxel grid of truncated distance function values and smoothed density value representations, respectively. Ppf-Net and its extension \cite{deng2018ppf, deng2018ppfnet} directly operate on unordered point sets to describe 3D keypoints. However, such methods require point cloud patches as input, resulting in an efficiency problem. This constraint severely limits its practicability, especially when fine-grained applications are needed. Besides these, dense feature description with a fully convolutional setting is proposed in FCGF \cite{choy2019fully}. For the detector learning, USIP \cite{li2019usip} utilizes a probabilistic chamfer loss to 
detect and localize keypoints in an unsupervised manner. Motivated by this, 3DFeat-Net \cite{yew20183dfeat} is the first attempt for 3D keypoints joint description and detection on point patches, which is then improved by D3Feat \cite{bai2020d3feat} to process full-frame point sets.
\vspace{-0.15cm}
\subsection{2D-3D Local Features Description}
\begin{figure*}
    \centering
    \includegraphics[width=0.8\textwidth]{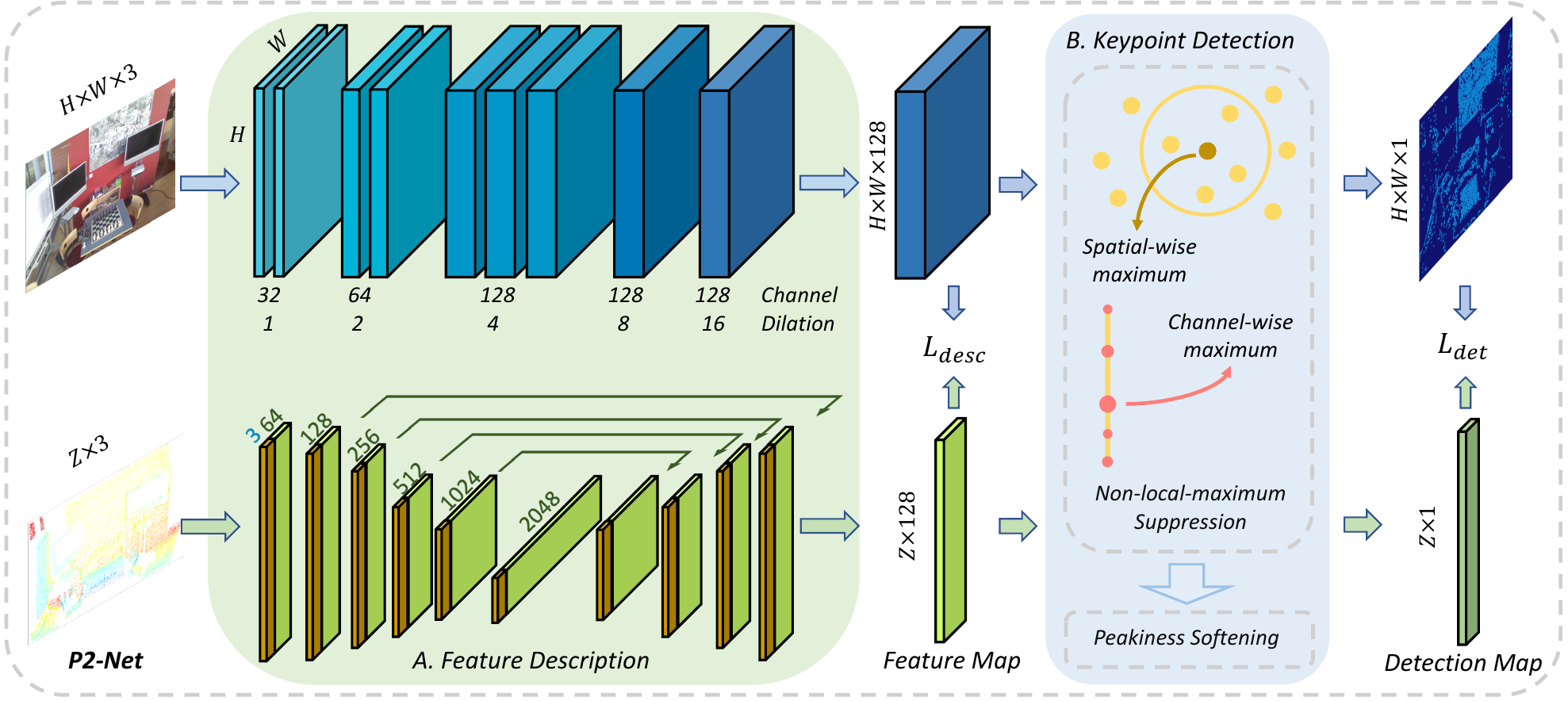}
        		    \vspace{-0.2cm}
    \caption{\textbf{An overview of the proposed P2-Net framework.} \wb{Our architecture is a two-branch fully convolutional network for the simultaneous 2D and 3D feature description (A) and keypoint detection (B). Such a network is jointly optimized with a descriptor loss $\mathcal{L}_{desc}$ enforcing the similarity of corresponding representations as well as a detector loss $\mathcal{L}_{det}$ encouraging higher detection scores for discriminative correspondences.}}
    \label{fig:framework}
    		    \vspace{-0.5cm}
\end{figure*}
\vspace{-0.15cm}
Unlike the well-researched area of learning descriptors in either a single 2D or 3D domain, little attention has been \wb{paid} on the learning of 2D-3D feature description. A 2D-3D descriptor is generated for object-level retrieval task by directly binding the hand-crafted 3D descriptor to a learned image descriptor \cite{li2015joint}. Similarly, 3DTNet \cite{xing20183dtnet} learns distinctive 3D descriptors for 3D patches with auxiliary 2D features extracted from 2D patches. Recently, both 2D3DMatch-Net \cite{feng20192d3d} and LCD \cite{pham2020lcd} propose to learn descriptors that allow direct matching across 2D and 3D local patches for retrieval problems. However, all these methods are patch-based, which are \wb{not applicable in real usages requiring high-resolution outputs.}
%as discussed in Section \ref{sec: intro}.
In contrast, we aim to extract per-point descriptors and detect keypoint locations in a single forward pass for efficient usage. 
%To the best of our knowledge, we are the first learning approach to achieve pixel-point level 2D-3D matching.
\vspace{-0.2cm}
\section{Pixel and Point Matching}
\vspace{-0.2cm}
In this section, we firstly introduce the architecture of the proposed \textbf{P2-Net} in detail, including feature description and keypoint detection. Next, we present our designed \textbf{P2-Loss}, composed of a circle-guided descriptor loss and a batch-hard detector loss. Finally, implementation details for both training and testing stages are provided.

\vspace{-0.15cm}
\subsection{P2-Net Architecture}
\vspace{-0.25cm}
% \subsection{Feature Description}
\PAR{Feature Description.}
The first step of our method is to obtain a 3D feature map $F^I\in\mathbb{R}^{H\times W\times C}$ from image $I$ and a 2D feature map $F^P\in\mathbb{R}^{Z\times C}$ from point cloud $P$, where $H$$\times$$W$ is the spatial resolution of the image, $Z$ is the number of points and $C$ is the dimension of the descriptors. Thus, the descriptor $d$ associated with the pixel $X$ and point $Y$ can be denoted as $d_{X}$ and $d_Y$, respectively,
\vspace{-0.25cm}
\begin{equation}\small
d_{X}=F^I_{X},\enspace d_Y=F^P_{Y}, d\in\mathbb{R}^C \enspace. 
\vspace{-0.25cm}
\end{equation}
%where $F^I_{hw}$ denotes the $h$-th row, $w$-th column of three-dimensional image feature map $F^I$ and $F^P_z$ denotes the $z$-th row of two-dimensional point cloud feature map $F^P$. 
After being L2-normalized to unit length, these descriptors can be readily compared between images and point clouds to establish correspondences using the cosine similarity as a metric. During training, the descriptors will be optimized so that a pixel and point pair in the scene produces similar descriptors, even when the image or point cloud contains strong changes or noise. For clarity, we still use $d$ to represent its normalized form in the following text.

\begin{figure}[h]
\vspace{-0.3cm}
    \centering
    \includegraphics[width=0.9\columnwidth]{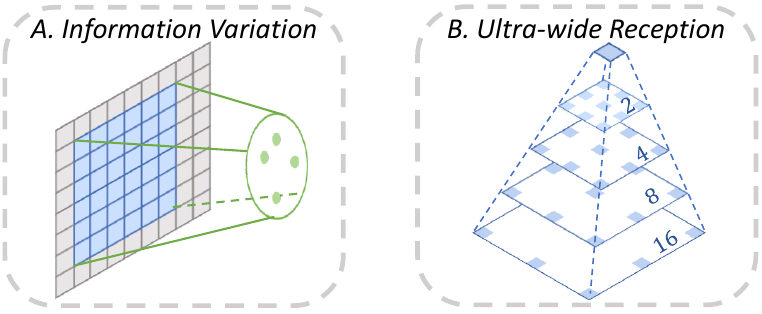}
    \vspace{-0.225cm}
    \caption{To mitigate the intrinsic information variation (A) between 2D and 3D local regions, an ultra-wide reception mechanism (B) with progressively doubling dilation values, up to 16, is applied in the 2D branch of feature description.}
    \label{fig:ultra}
    		    \vspace{-0.425cm}
\end{figure}

As shown in Fig.~\ref{fig:framework}.A, two fully convolutional networks are exploited to separately perform feature description on images and point clouds. However, properly associating pixels with points through descriptors is non-trivial because of the intrinsic variation in information density between 2D and 3D local regions (Fig.~\ref{fig:ultra}.A). Specifically, the local information represented by a point is typically larger than a pixel due to the sparsity of point clouds. To address the issue of association on asymmetrical embeddings and better capture the local geometry information, we design the 2D extractor based on an ultra-wide reception mechanism, shown in Fig.~\ref{fig:ultra}.B. For computational efficiency, such a mechanism is achieved through nine $3 \times 3$ convolutional layers with progressively doubling dilation values, from 1 to 16. Finally, a $H$$\times$$W$$\times$$128$ feature map is generated and then its corresponding $H$$\times$$W$$\times$$1$ detection map can be computed. In a similar vein, we modify KPconv \cite{thomas2019kpconv} to output a 128D descriptor and a score for each point in the input point cloud.

% \subsection{Keypoint Detection}
\PAR{Keypoint Detection.}
As illustrated in Fig.~\ref{fig:framework}.B, we determine keypoints by performing a \wb{peakiness-softened non-local-maximum suppression} \cite{luo2020aslfeat} across the spatial and channel dimensions of a feature map. 
% Similar to \cite{bai2020d3feat}, we utilize peakiness score \cite{luo2020aslfeat, bai2020d3feat} as a keypoint measurement to avoid density-invariant problem of point cloud when evaluate the local maximum.
Given a feature map $F\in\mathbb{R}^{T \times C}$, where $T$$=$$H$$\times$$W$ for images and $T$$=$$Z$ for point clouds.
% donates the shape of detection map (${T=H\times W}$ for image and ${T=Z}$ for point cloud):
The requirement for a pixel or point $\rho_{t}$ to be detected by a non-local-maximum suppression is
\wb{
    \vspace{-0.25cm}
    \begin{equation}\small
        \begin{aligned}
        \rho_{t}\text{ is a detection } \Longleftrightarrow & \ F^c_{t} \text{ is a local max in } F^c_{\mathcal{R}_{\rho_{t}}} \\
        & \text{with } c = \arg\max_{k} \ F_t^k \enspace,  
        \label{hard detection}
        \end{aligned}
            \vspace{-0.25cm}
    \end{equation}
in which $F^c_{t}$ represents the feature response at the position $t$ and channel $c$. $\mathcal{R}_{\rho_{t}}$ denotes the neighborhood of $\rho_{t}$}. 

\wb{During training, the above procedure is softened to be trainable and density-invariant using peakiness \cite{revaud2019r2d2}:
\vspace{-0.3cm}
\begin{equation}\small
\vspace{-0.3cm}
    \begin{split}
    &\alpha^c_{t} = \text{softplus}(F^c_{t} - \frac{1}{|\mathcal{R}_{\rho_{t}}|}\sum_{\rho_{t^\prime}\in\mathcal{R}_{\rho_{t}}} F^c_{t^\prime}) \enspace, \\
    &\beta^c_{t}  = \text{softplus}(F^c_{t} -  \frac{1}{C}\sum_{k} F^k_{t}) \enspace ,
    \end{split}
\end{equation}
where $\alpha$ and $\beta$ are the spatial-wise and channel-wise detection scores, respectively. The final keypoint detection score of $\rho_{t}$ that takes both criteria into account is:}
\vspace{-0.2cm}
\begin{equation}\small
    \xi_{\rho_{t}} = \max_c \left( \alpha^c_{t} \beta^c_{t} \right) \enspace .
    \vspace{-0.2cm}
     \label{eq:softscore}
\end{equation}
\noindent \wb{During testing, pixels or points with top scores will be selected as keypoints for matching.}
\begin{figure}
    \centering
    \includegraphics[width=0.75\columnwidth]{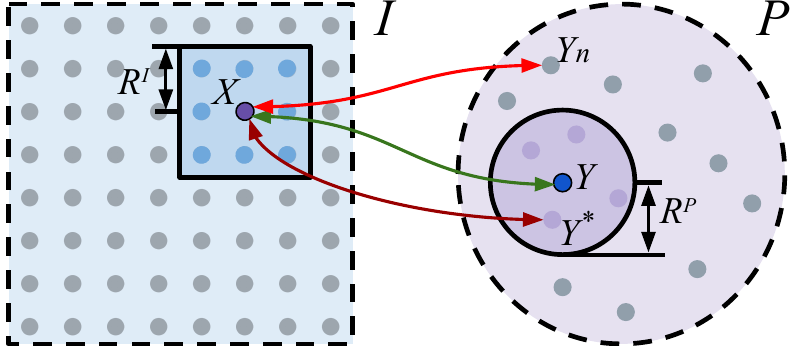}
    \vspace{-0.2cm}
    \caption{\textbf{Definitions of pixel-point pairs.} In a pair of image $I$ and point cloud $P$, \bm{$X$}$\color{Green4}{\bm{\leftrightarrow}}$\bm{$Y$} is a correspondence (pixel $X$$\in$$I$ and point $Y$$\in$$P$). From the image perspective, \bm{$X$}$\color{red}{\bm{\leftrightarrow}}$\bm{$Y_n$} demonstrates a negative match where $Y_n$ lies outside $R^P$ (the neighborhood of $Y$), denoting a negative point of $X$. \bm{$X$}$\color{DarkRed}{\bm{\leftrightarrow}}$\bm{$Y^*$} represents the hardest negative match and $Y^*$ is the hardest negative point of $X$ in the whole point cloud space. The negative and the hardest negative matches in the perspective of a point cloud are the opposite.}
    \label{fig:desc}
    		    \vspace{-0.4cm}
\end{figure}
\vspace{-0.1cm}
\wb{\subsection{P2-Loss Formulation} \label{sec:loss}}
\vspace{-0.1cm}
To make the proposed network describe and detect 2D and 3D keypoints in a single forward pass, we design a novel loss that jointly optimizes the description and detection objectives for both pixels and points, named \textbf{P2-Loss}:
\vspace{-0.3cm}
\begin{equation}\small
\vspace{-0.1cm}
\label{eq:training loss}
    \mathcal{L}_{P2} = \mathcal{L}_{desc} + \lambda \mathcal{L}_{det} \enspace.
\end{equation}
It consists of a circle-guided descriptor loss $\mathcal{L}_{desc}$ that expects distinctive descriptors to avoid incorrect match assignments, a batch-hard detector loss $\mathcal{L}_{det}$ that encourages keypoints to be repeatable under viewpoint or illumination changes, and a balance factor $\lambda$ between them.

\PAR{Circle-guided Descriptor Loss.}
To learn distinctive descriptors, various optimization strategies like hard-triplet and hard-contrastive losses \cite{dusmanu2019d2, luo2020aslfeat, bai2020d3feat} have been widely used in 2D or 3D domain. However, these formulations only focus on hard negative matches, and experimentally we found that they did not converge in our 2D-3D context. Inspired by the Circle Loss~\cite{sun2020circle} using weighting factors and the circular decision boundary, \wb{we design a \textbf{circle-guided descriptor} loss with a full sampling strategy instead of only considering the hard negative matches, which allows self-paced optimization and avoids convergence ambiguity.}

Given a correspondence \bm{$X$}$\color{Green4}{\bm{\leftrightarrow}}$\bm{$Y$} between image $I$ and point cloud $P$ in Fig.~\ref{fig:desc}, we can define a positive cosine similarity $s_{p}$ for corresponding descriptors $d_X$ and $d_Y$ as:
\vspace{-0.25cm}
\begin{equation}\small
    \vspace{-0.15cm}
    s_{p} = d_X d_Y = \sum_{c} d^c_X d^c_Y,
\end{equation}
% \noindent where the negative samples $d_{X^N}$ and $d_{Y^N}$ are the negatives that lie outside of a local neighbourhood $N$ of the correct correspondence.
From the view of image, we fully sample negative pairs \bm{$X$}$\color{red}{\bm{\leftrightarrow}}$\bm{$Y_n$} and define a negative cosine similarity set $s^I_{n}$ for all negative descriptor pairs $d_X$ and $d_{Y^{1,\cdots,j}_n}$ as:
\vspace{-0.2cm}
\begin{equation}\small
\vspace{-0.2cm}
    s^I_{n} = \left\{d_X d_{Y^1_n},\cdots,d_X d_{Y^j_n}\right\},\enspace \text{s.t. }\lVert Y^j_n-Y \rVert_2 \textgreater R^P,
\end{equation}
where $Y^j_n$ denotes a negative sample of pixel $X$ lying outside $R^P$ which is the safe radius of point $Y$. The circle-guided descriptor loss of the image part is then derived as:
\vspace{-0.3cm}
\begin{equation}\small
\vspace{-0.10cm}
 \begin{split}
    &\mathcal{L}^I_{desc} = \frac{1}{|\mathcal{C}|} \sum_{i} \log \Big [ 1 + e^{ \eta^i_p (1 - m - s_p^i)}
\sum_{j} e^{\eta^j_n (s_n^{I^j_i}-m)}\Big ],
    \end{split}
\label{eq:descriptor}
\end{equation}
in which $\mathcal{C}$ is the set of correspondences between image $I$ and point cloud $P$ used for optimization in each step, $\eta^i_p$=$\zeta(1$+$m$-$s_p^i)$ and $\eta^j_n$=$\zeta(s_n^{I^j_i}$+$m)$ represent weighting factors with a scale factor $\zeta$ that expect $s_p^i$\textgreater$1$-$m$ and $s_n^{I^j_i}$\textless$m$ in a self-paced manner. \wb{With that said, the margin $m$ controls the radius of the circular decision boundary at $(s_n^{I^j_i}$-$0)^2 $+$ (s_p^i$-$1)^2$=$2m^2$.} The reverse loss $\mathcal{L}^P_{desc}$ for point cloud $P$ is calculated 
in the same way for a total circle-guided descriptor loss
$\mathcal{L}_{desc}=\frac{1}{2}(\mathcal{L}^I_{desc} + \mathcal{L}^P_{desc})$. 

\PAR{Batch-hard Detector Loss.}
In the case of detection, keypoints should be sufficiently distinctive to be repeatably detected. Achieving this objective, however, faces two practical challenges: \textbf{1)} the ultra-wide reception mechanism in feature description may leave spatially close pixels possessing very similar descriptors;  \textbf{2)} the full sampling strategy in our descriptor loss is only effective to negative matches outside a safe region. Both of them will reduce the distinctiveness of keypoints and thus cause erroneous assignments. To this end, we design a \textbf{batch-hard detector} loss with applying \textit{hardest-in-batch} strategy~\cite{mishchuk2017working} on the whole image or point cloud space but not on a specific area, encouraging optimal distinctiveness and repeatability.

\wb{Similar to the hardest negative match \bm{$X$}$\color{DarkRed}{\bm{\leftrightarrow}}$\bm{$Y^*$} in Fig.~\ref{fig:desc}, $Y^*_i$ is determined by $\argmax_{Y^l_i \neq Y_i} (d_{Y^l_i} d_{X_i})$ and denotes the hardest negative point of $X_i$ in the whole point cloud space. In extension to $X^*_i$, we can thus define the hardest negative similarity $s^i_{n^*}$ as $\max(d_{X_i} d_{Y^*_i}, d_{Y_i} d_{X^*_i})$. Additionally, $\xi_{X_i}$ and $\xi_{Y_i}$ are the soft detection scores at pixel $X_i$ and point $Y_i$, respectively. With above definitions, we then formualte the batch-hard detector loss as:}
\vspace{-0.2cm}
\begin{equation}\small
\vspace{-0.2cm}
 \begin{split}
    &\mathcal{L}_{det} = \sum_{i \in \mathcal{C}} \frac{\xi_{X_i} \xi_{Y_i}}{\sum\limits_{q\in \mathcal{C}} \xi_{X_q}\xi_{Y_q}} (s^i_{n^*} - s^i_{p}),
\end{split}
\label{eq:detector}
\end{equation}
Intuitively, such a detector loss seeks for higher detection scores for more discriminative correspondences. Specifically, $L_{det}$ expects $\xi_{X_i}$ and $\xi_{Y_i}$ to be high if $s^i_{n^*}$\textless$s^i_{p}$. Moreover, the more discriminative correspondences, with a lower value of ($s^i_{n^*}-s^i_{p}$), are encouraged to possess higher relative detection scores and vice-versa.
\vspace{0.3cm}
\subsection{Implementation Details}
\vspace{-0.1cm}
\PAR{Training.}
We implement our approach with PyTorch. During the training, we use a batch size of 1 and all image-point cloud pairs with more than 128 pixel-point correspondences. For the sake of computational efficiency, $|\mathcal{C}|$=$128$ correspondences are randomly sampled from each pair to optimize in each step. We set the balance factor $\lambda$=$1$, the margin {$m$=$0.2$}, scale factor {$\zeta$=$10$}, image neighbour $R_{I}$=$12$ pixels, point cloud neighbour $R_{P}$=$0.015$ m. Finally, we train the network with the ADAM solver and use an initial learning rate of $10^{-4}$ with exponential decay\textsuperscript{\ref{app:details}}.
% Please see the supplementary material for additional implementation details.
\vspace{-0.0cm}
\PAR{Testing.}
During testing, we exploit the hard selection strategy demonstrated in Eq.~\ref{hard detection} rather than soft selection to mask detections that are spatially too close. Additionally, the SIFT-like edge elimination is applied for image keypoints detection. For evaluation, we select the top-K keypoints corresponding to the detection scores calculated in Eq.~\ref{eq:softscore}.

\section{Experiments}

% In our experiments, 
We first demonstrate the effectiveness of proposed P2-Net on the \textit{direct} pixel and point matching task, and then evaluate it on a downstream task, namely visual localization. Furthermore, we examine the generalization ability of our designed P2-Loss in single 2D and 3D domains, by comparing with the state-of-the-art methods in both image matching and point cloud registration tasks respectively. %Finally, we evaluate our loss in the homogeneous matching tasks of 2D image matching and 3D point cloud registration, to demonstrate the generalization ability of our proposed loss.
Finally, we investigate the effect of the loss selection.
\vspace{-0.0cm}
\subsection{Image and Point Cloud Matching}
\vspace{-0.0cm}
To achieve fine-grained image and point cloud matching, a dataset of image and point cloud pairs annotated with pixel and point correspondences is required. To the best of our knowledge, there is no publicly available dataset with such correspondence labels. To address this issue, we annotated the 2D-3D correspondence labels\footnote{Please refer to the supplementary material for more details.\label{app:details}} on existing 3D datasets containing RGB-D scans. Specifically, the 2D-3D correspondences of our dataset are generated on the 7Scenes dataset \cite{glocker2013real, shotton2013scene}, consisting of seven indoor scenes with 46 RGB-D sequences recorded under various camera motion status and different conditions, e.g. motion blur, perceptual aliasing and textureless features in the room. These conditions are widely known to be challenging for both image and point cloud matching.
\vspace{-0.3cm}
\subsubsection{Evaluation on Feature Matching}\label{sec:matching}
\vspace{-0.1cm}
We adopt the same data splitting strategy for the 7Scenes dataset as in \cite{glocker2013real, shotton2013scene} to prepare the training and testing set. Specifically, 18 sequences are selected for testing, which contain partially overlapped image and point cloud pairs, and the ground-truth transformation matrices.
\vspace{-0.0cm}
\PAR{Evaluation metrics.} To comprehensively evaluate the performance of our proposed P2-Net  and P2-Loss on fine-grained image and point cloud matching, five metrics widely used in previous image or point cloud matching tasks \cite{luo2020aslfeat, dusmanu2019d2, balntas2017hpatches, li2019usip, zeng20173dmatch, dong2020registration, bai2020d3feat} are adopted: 1) Feature Matching Recall, the percentage of image and point cloud pairs with the inlier ratio above a threshold ($\tau_1=0.5$); 2) Inlier Ratio, the percentage of correct pixel-point matches over all possible matches, where a correct match is accepted if the distance between the pixel and point pair is below a threshold ($\tau_2=4.5$cm) under its ground truth transformation; 3) Keypoint Repeatability, the percentage of repeatable keypoints over all detected keypoints, where a keypoint in the image is considered repeatable if its distance to the nearest keypoint in the point cloud is less than a threshold ($\tau_3=2$cm) under the true transformation; 4) Recall, the percentage of correct matches over all ground truth matches; 5) Registration Recall, the percentage of image and point cloud pairs with the estimated transformation error smaller than a threshold (RMSE$<5$cm)\textsuperscript{\ref{app:details}}.

\begin{table}[t]
\Huge
    \centering
    \resizebox{\linewidth}{!}{
\begin{tabular}{lccccccc}
\textbf{\# Scenes} & \textbf{Chess} & \textbf{Fire} & \textbf{Heads} & \multicolumn{1}{l}{\textbf{Office}} & \textbf{Pumpkin} & \textbf{Kitchen} & \multicolumn{1}{l}{\textbf{Stairs}} \\ \hline
\multicolumn{8}{c}{\textit{Feature Matching Recall}}                                                                                                                                   \\ \hline
SIFT + SIFT3D              & \multicolumn{7}{c}{Not Match}                                                                                                                                     \\
\wb{P2[D2\_Triplet]}       & \multicolumn{7}{c}{Not Converge}                                                                                                                                  \\
\wb{P2[D3\_Contrastive]}   & \multicolumn{7}{c}{Not Converge}                                                                                                                                  \\
\wb{P2[R2D2]}      & 95.1           & 97.3          & \textbf{100}   & 89.4                                & 91.1             & 88.7             & 16.2                                \\
\wb{P2[ASL]}       & 95.3           & 96.0          & \textbf{100}   & 34.3                                & 41.6             & 47.5             & 11.9                                \\
P2[w/o Det]        & 93.0           & 97.0          & 99.1           & 73.8                                & 61.5             & 43.8             & 15.0                                \\
P2[Mixed]          & 92.5           & 96.0          & 99.7           & 74.6                                & 52.2             & 69.0             & 15.8                                \\
P2[D2\_Det]        & \textbf{100}   & 99.7          & \textbf{100}   & 93.6                                & 98.4             & 94.0             & 74.3                                \\
P2[D3\_Det]        & 99.0           & 99.7          & \textbf{100}   & 83.8                                & 68.0             & 78.4             & 17.8                                \\
P2[Rand]           & \textbf{100}   & 99.6          & 99.8           & 90.8                                & 83.2             & 82.5             & 14.3                                \\
P2[Full]             & \textbf{100}   & \textbf{100}  & \textbf{100}   & \textbf{97.3}                       & \textbf{98.5}    & \textbf{96.3}    & \textbf{88.8}                       \\ \hline
\multicolumn{8}{c}{\textit{Registration Recall}}                                                                                                                                       \\ \hline
\wb{P2[R2D2]}      & 81.0           & 78.5          & 73.1           & 79.7                                & 75.6             & 77.1             & 60.8                                \\
\wb{P2[ASL]}       & 70.5           & 66.0          & 63.4           & 52.9                                & 41.6             & 48.0             & 38.2                                \\
P2[w/o Det]        & 68.0           & 64.5          & 53.8           & 59.6                                & 48.4             & 56.1             & 42.3                                \\
P2[Mixed]          & 72.5           & 66.5          & 20.9           & 59.1                                & 53.2             & 63.5             & 25.6                                \\
P2[D2\_Det]        & 86.0           & 75.5          & 74.2           & 70.8                                & 80.0             & 74.3             & 78.3                                \\
P2[D3\_Det]        & 80.5           & 70.0          & 81.7           & 76.3                                & 65.5             & 70.6             & 70.9                                \\
P2[Rand]           & 86.5           & 81.5          & 82.6           & 78.9                                & 75.5             & 77.2             & 74.3                                \\
P2[Full]             & \textbf{87.0}  & \textbf{82.4} & \textbf{84.5}  & \textbf{83.4}                       & \textbf{88.7}    & \textbf{82.7}    & \textbf{82.6}                       \\ \hline
\multicolumn{8}{c}{\textit{Keypoint Repeatability}}                                                                                                                                    \\ \hline
\wb{P2[R2D2]}      & 36.6           & 40.3          & 45.2           & 33.4                                & 30.3             & 32.1             & 33.1                                \\
\wb{P2[ASL]}       & 18.7           & 19.2          & 33.8           & 13.8                                & 12.9             & 15.5             & 11.9                                \\
P2[w/o Det]        & 17.4           & 17.8          & 37.0           & 18.2                                & 16.0             & 15.7             & 17.7                                \\
P2[Mixed]          & 23.3           & 26.6          & 26.0           & 30.0                                & 29.9             & 31.3             & 24.7                                \\
P2[D2\_Det]        & 41.7           & 39.8          & 40.6           & 34.8                                & 32.7             & 31.6             & 34.9                                \\
P2[D3\_Det]        & 24.9           & 21.8          & 38.1           & 24.5                                & 19.6             & 23.8             & 21.8                                \\
P2[Rand]           & 36.1           & 37.0          & 46.1           & 33.5                                & 30.4             & 32.2             & 36.1                                \\
P2[Full]             & \textbf{50.4}  & \textbf{47.1} & \textbf{50.2}  & \textbf{38.0}                       & \textbf{45.2}    & \textbf{38.3}    & \textbf{48.1}                       \\ \hline
\multicolumn{8}{c}{\textit{Recall}}                                                                                                                                                    \\ \hline
\wb{P2[R2D2]}      & 28.5           & 26.7          & 24.7           & 25.0                                & 24.6             & 26.4             & 16.0                                \\
\wb{P2[ASL]}       & 28.8           & 26.3          & 16.5           & 21.7                                & 21.4             & 23.8             & 13.8                                \\
P2[w/o Det]        & 29.1           & 26.9          & 23.1           & 25.3                                & 22.0             & 23.8             & 14.4                                \\
P2[Mixed]          & 30.1           & 26.2          & 25.2           & 24.5                                & 24.1             & 26.9             & 15.1                                \\
P2[D2\_Det]        & 30.3           & 28.9          & 26.1           & 27.0                                & \textbf{29.6}    & 28.7             & 17.7                                \\
P2[D3\_Det]        & 31.8           & 31.1          & 26.4           & 26.6                                & 25.6             & 27.5             & 17.1                                \\
P2[Rand]           & 31.4           & 30.8          & 25.7           & 29.5                                & 28.0             & 30.6             & 17.6                                \\
P2[Full]             & \textbf{32.7}  & \textbf{33.7} & \textbf{26.6}  & \textbf{30.6}                       & \textbf{29.6}    & \textbf{32.3}    & \textbf{20.1}                       \\ \hline
\multicolumn{8}{c}{\textit{Inlier Ratio}}                                                                                                                                              \\ \hline
\wb{P2[R2D2]}      & 65.5           & 66.5          & 69.8           & 54.0                                & 54.5             & 55.3             & 38.5                                \\
\wb{P2[ASL]}       & 55.9           & 60.8          & 64.9           & 44.7                                & 45.7             & 47.6             & 34.2                                \\
P2[w/o Det]        & 52.7           & 56.3          & 71.0           & 46.1                                & 47.3             & 49.9             & 36.2                                \\
P2[Mixed]          & 51.5           & 55.2          & 67.4           & 52.1                                & 50.1             & 56.7             & 35.1                                \\
P2[D2\_Det]        & 68.2           & 72.2          & 74.9           & 58.0                                & \textbf{61.4}    & 59.3             & 42.9                                \\
P2[D3\_Det]        & 61.1           & 64.6          & 75.4           & 51.3                                & 47.6             & 51.8             & 37.9                                \\
P2[Rand]           & 58.5           & 61.4          & 76.2           & 53.2                                & 50.0             & 53.4             & 40.4                                \\
P2[Full]             & \textbf{73.9}  & \textbf{76.0} & \textbf{77.4}  & \textbf{60.3}                       & 60.8             & \textbf{65.2}    & \textbf{45.2}                      
\end{tabular}
}
			    \vspace{-0.2cm}
    \caption{\textbf{Comparisons on the 7Scenes dataset \cite{glocker2013real, shotton2013scene}}. Evaluation metrics are reported within given thresholds.}
	\label{tbl:7scenes}
			    \vspace{-0.5cm}
\end{table}
% Feature Matching Recall, Registration Recall, Keypoint Repeatability, Recall and Inlier Ratio are reported within given thresholds.

\vspace{-0.0cm}
\PAR{Comparisons on descriptors and networks.}
To study the effects of descriptors, we report the results of 1) traditional SIFT and SIFT3D descriptors; 2) P2-Net trained with the D2-Net loss (P2[D2\_Triplet]) \cite{dusmanu2019d2} and 3) P2-Net trained with the D3Feat loss (P2[D3\_Contrastive]) \cite{bai2020d3feat}. Besides, to demonstrate the superiority of the 2D branch in P2-Net, we replace it with 4) the R2D2 network (P2[R2D2]) \cite{revaud2019r2d2} and 5) the ASL network (P2[ASL]) \cite{luo2020aslfeat}. Other training or testing settings are the same with the proposed architecture trained with our proposed loss (P2[Full]) for a fair comparison. \wb{Among them, both P2[R2D2] and P2[Full] adopt L2-Net-style ~\cite{tian2017l2} 2D feature extractors but the latter is improved by our ultra-wide reception mechanism}.

As shown in Tab.~\ref{tbl:7scenes}, traditional descriptors fail to be matched, as hand-designed 2D and 3D descriptors are heterogeneous. Both P2[D2\_Triplet] and P2[D3\_Contrastive] are not able to guarantee convergence on the pixel-point matching task. However, when adopting our loss, P2[R2D2] and P2[ASL] models not only converge but also present promising performance in most scenes, except the challenging Stairs scene, due to the intrinsic feature extractor limitation of R2D2 and ASL. \wb{Moreover, the comparison between P2[R2D2] and P2[Full] also demonstrates the effectiveness of the ultra-wide reception mechanism}. Overall, our P2[Full] performs consistently better regarding all evaluation metrics, outperforming all competitive methods by a large margin on all scenes. 

\vspace{-0.0cm}
\PAR{Comparisons on detectors.}
In order to demonstrate the importance of jointly learning the detector and descriptor, we report the results of P2-Net trained with our circle-guided descriptor loss and :1) without a detector but with randomly sampled keypoints (P2[w/o Det]); 2) without a detector but with conventional SIFT and SIFT3D keypoints (P2[Mixed]); 3) with the original D2-Net detector (P2[D2\_Det]) \cite{dusmanu2019d2}; 4) with the D3Feat detector (P2[D3\_Det]) \cite {bai2020d3feat}; 5) with our batch-hard detector loss but using randomly sampled keypoints  (P2[Rand]) for testing to indicate the superiority of our proposed detector. 

As can be seen from Tab.~\ref{tbl:7scenes}, when a detector is not jointly trained with entire model, P2[w/o Det] shows the worst performance on all evaluation metrics and scenes. Such indicators are slightly improved by P2[Mixed] after introducing traditional detectors. Nevertheless, when the proposed detector is used, P2[Rand] achieves better results than P2[Mixed]. These results conclusively indicate that a joint learning with detector is also advantageous to strengthening the descriptor learning itself. Similar improvements can also be observed in both P2[D2\_Det] and P2[D3\_Det]. Clearly, our P2[Full] is able to maintain a competitive matching quality in terms of all evaluation metrics, if our loss is fully enabled. It is worth mentioning that, particularly in the scene of Stairs, P2[Full] is the only method that achieves outstanding matching performance on all metrics. In contrast, most of the other competing methods fail due to the highly repetitive texture in this challenging scenario. It indicates that the keypoints are robustly detected and matched even under challenging condition, which is a desired property for reliable keypoints to possess\footnote{Please refer to the supplementary material for additional results\label{app:results}.}.

%To study the effect of detector, we comprehensively compare different types of detectors. To demonstrate the benefit from a joint learning of two tasks, we use only the descriptors loss to train a model without the detector learning .
\vspace{-0.0cm}
\PAR{Qualitative results.}
Fig.~\ref{fig:visualization} shows the top-1000 detected keypoints for images and point clouds from different scenes. Detected pixels from images (left, green) and detected points from point cloud (right, red) are displayed on Chess and Stairs. For clarity, we randomly highlight some of good matches (blue, orange) to enable better demonstration of the correspondence relations. As can be seen, by our proposed descriptors, such detected pixels and points are directly and robustly associated, which is essential for real-world downstream applications (e.g., cross-domain information retrieval and localization tasks). Moreover, as our network is jointly trained with the detector, the association is able to bypass regions that cannot be accurately matched, such as the repetitive patterns. More specifically, our detectors mainly focus on the geometrically meaningful areas (e.g. object corners and edges) rather than the feature-less regions (e.g. floors, screens and tabletops), and thus show better consistency over environmental changes\textsuperscript{\ref{app:results}}.

\vspace{-0.4cm}
\subsubsection{Application on Visual Localization} 
\vspace{-0.2cm}
\begin{figure}
    \centering
    \includegraphics[width=0.92\columnwidth]{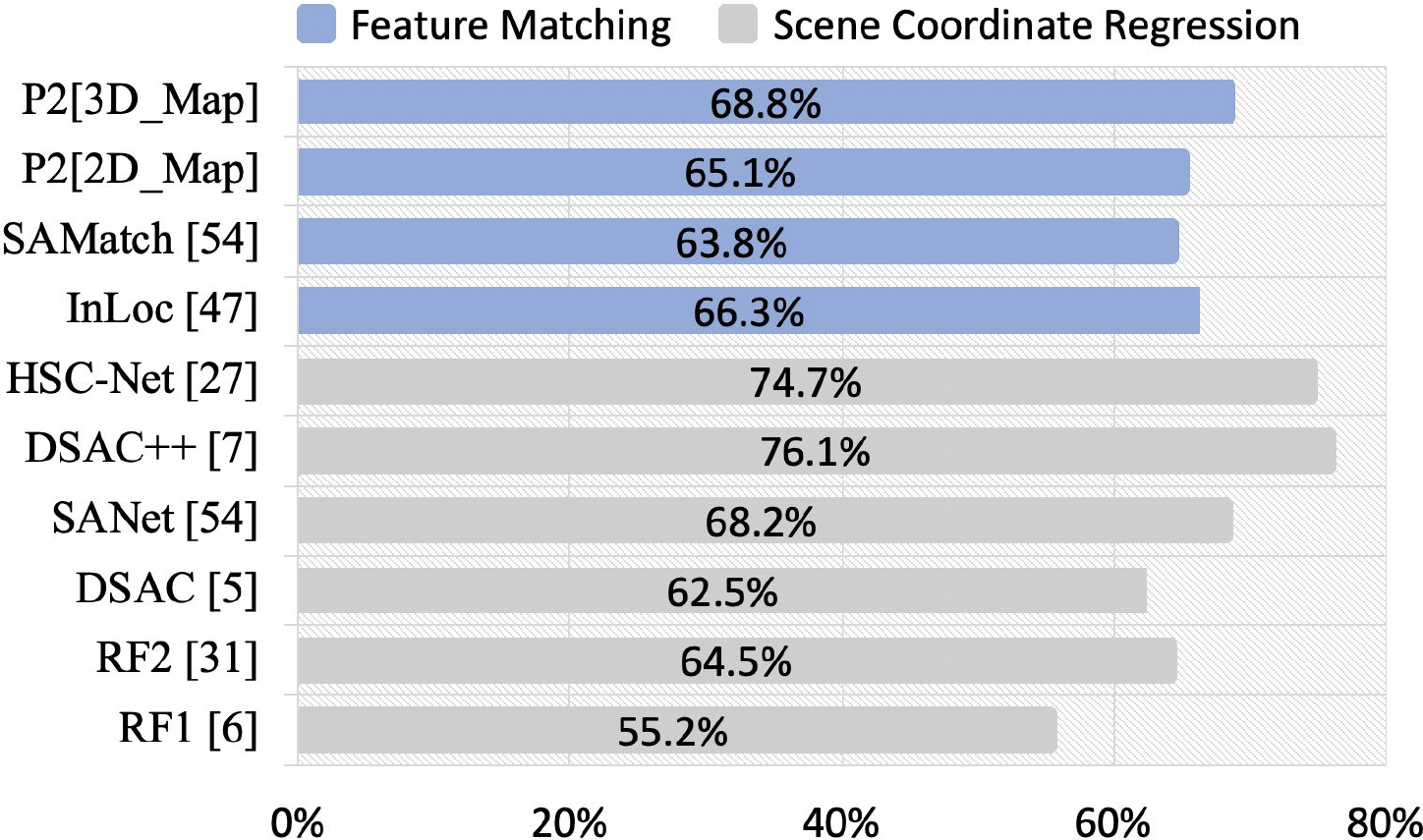}
    			    \vspace{-0.2cm}
    \caption{\textbf{Comparisons on visual localization.} Percentage of estimated camera poses falling within ($5 cm$, $5 \degree$).}
    \label{fig:visual localization}
    		    \vspace{-0.5cm}
\end{figure}
% by RF1 \cite{brachmann2016uncertainty}, RF2 \cite{massiceti2017random}, DSAC \cite{brachmann2017dsac}, DSAC++ \cite{brachmann2018learning}, InLoc \cite{taira2018inloc}, SAMatch (matching-based SANet) \cite{yang2019sanet}, SANet \cite{yang2019sanet}, HSC-Net \cite{li2020hierarchical} and our approaches
To further illustrate the practical usage of P2-Net, we perform a downstream task of visual localization \cite{wang2020atloc, li2020hierarchical} on the 7Scenes dataset. The key localization challenge here lies in the fine-grained matching between pixels and points under significant motion blur, perceptual aliasing and textureless patterns. We evaluate  our method against the 2D feature matching based \cite{taira2018inloc, yang2019sanet} and scene coordinate regression pipelines \cite{brachmann2016uncertainty, massiceti2017random, brachmann2017dsac, brachmann2018learning, yang2019sanet, li2020hierarchical}. \textit{Note that existing baselines are only able to localize queried images in 3D maps, while our method is not limited by this but can localize reverse queries from 3D to 2D as well.} The following experiments are conducted to show the uniqueness of our method: 1) recovering the camera pose of a query image in a given 3D map (P2[3D\_Map]) and 2) recovering the pose of a query point cloud in a given 2D map (P2[2D\_Map]). 

\vspace{-0.05cm}
\PAR{Evaluation protocols.} We follow the same evaluation pipeline used in \cite{sattler2016efficient, taira2018inloc, yang2019sanet}. This pipeline typically takes input as query images and a 3D point cloud submap (e.g., retrieved by NetVLAD \cite{arandjelovic2016netvlad}), and utilizes traditional hand-crafted or pre-trained deep descriptors to establish the matches between pixels and points. Such matches are then taken as the input of PnP with Ransac \cite{brachmann2017dsac} to recover the final camera pose. Here, we adopt the same setting in \cite{yang2019sanet} to construct the 2D or 3D submaps that cover a range up to $49.6$ cm. Recall that our goal is to evaluate the effects of matching quality for visual localization, we therefore assume the submap has been retrieved and focus more on comparing the distinctiveness of keypoints. During testing, we select the top $10,000$ detected pixels and points to generate matches for camera pose estimation.
\vspace{-0.05cm}
\PAR{Results.} We follow \cite{taira2018inloc, yang2019sanet} to evaluate models on $\frac{1}{10}$ testing frames. The localization accuracy is measured in terms of percentage of predicted poses falling within the threshold of ($5 cm$, $5 \degree$). As shown in Fig.~\ref{fig:visual localization}, when matching 2D features against 3D map, our P2[3D\_Map] ($68.8\%$), outperforms InLoc \cite{taira2018inloc} and SAMatch \cite{yang2019sanet} by $2.6\%$ and $5 \%$, respectively, where the conventional feature matching approach are used to localize query images. Moreover, our P2[3D\_Map] presents better results than most of the scene coordinated based methods such as RF1 \cite{brachmann2016uncertainty}, RF2\cite{massiceti2017random}, DSAC \cite{brachmann2017dsac} and SANet \cite{yang2019sanet}. DSAC++ \cite{brachmann2018learning} and HSC-Net ~\cite{li2020hierarchical} still show better performance than ours, because they are trained for individual scene specifically and use individual models for testing. In contrast, we directly use the single model trained from P2[Full] in Sec.~\ref{sec:matching}, which is scene agnostic. In the unique application scenario that localizes 3D queries in a 2D map, our P2[2D\_Map] also shows promising performance, reaching $65.1\%$. However, other baselines are not capable of realizing this inverse matching.
% \zp{How do you train our network?}
\vspace{-0.4cm}
\subsection{Matching under Single Domains}
\vspace{-0.2cm}
\newpage
In this experiment, we demonstrate how our novel proposed P2-Loss formulation can greatly improve the performance of state-of-the-art 2D and 3D matching networks.
\wb{
\begin{table}[t]
    \centering
     \resizebox{\linewidth}{!}{
\begin{tabular}{cl|l|l|c|l|c|lcl|c|lcl}
\multicolumn{1}{l}{\multirow{2}{*}{}}   & \multicolumn{3}{c||}{\multirow{2}{*}{}} & \multicolumn{2}{c|}{SP} & \multicolumn{4}{c|}{D2-Net \cite{dusmanu2019d2}} & \multicolumn{4}{c}{ASLFeat \cite{luo2020aslfeat}} \\
\multicolumn{1}{l}{}                    & \multicolumn{3}{c||}{}                  & \multicolumn{2}{c|}{\cite{detone2018superpoint}}   & \multicolumn{2}{c}{Triplet} & \multicolumn{2}{c|}{\textbf{Ours}} & \multicolumn{2}{c}{Contra} & \multicolumn{2}{c}{\textbf{Ours}}                 \\ \hline \hline
\multirow{3}{*}{Illum} & \multicolumn{3}{l||}{HEstimation}       & \multicolumn{2}{c|}{0.877}       & \multicolumn{2}{c}{0.818}          & \multicolumn{2}{c|}{\textbf{0.857}} & \multicolumn{2}{c|}{{\ul \textbf{0.919}}} & \multicolumn{2}{c}{0.915}                \\
                                                                          & \multicolumn{3}{l||}{Precision}         & \multicolumn{2}{c|}{0.629}       & \multicolumn{2}{c}{0.650}          & \multicolumn{2}{c|}{\textbf{0.664}} & \multicolumn{2}{c|}{0.774}                & \multicolumn{2}{c}{{\ul \textbf{0.787}}} \\
                                                                          & \multicolumn{3}{l||}{Recall}            & \multicolumn{2}{c|}{0.565}       & \multicolumn{2}{c}{\textbf{0.564}} & \multicolumn{2}{c|}{0.560}          & \multicolumn{2}{c|}{0.696}                & \multicolumn{2}{c}{{\ul \textbf{0.726}}} \\ \hline
\multirow{3}{*}{View}  & \multicolumn{3}{l||}{HEstimation}       & \multicolumn{2}{c|}{{\ul 0.651}} & \multicolumn{2}{c}{0.553}          & \multicolumn{2}{c|}{\textbf{0.581}} & \multicolumn{2}{c|}{0.542}                & \multicolumn{2}{c}{{\textbf{0.598}}} \\
                                                                          & \multicolumn{3}{l||}{Precision}         & \multicolumn{2}{c|}{0.595}       & \multicolumn{2}{c}{0.564}          & \multicolumn{2}{c|}{\textbf{0.576}} & \multicolumn{2}{c|}{0.708}                & \multicolumn{2}{c}{{\ul \textbf{0.740}}} \\
                                                                          & \multicolumn{3}{l||}{Recall}            & \multicolumn{2}{c|}{0.446}       & \multicolumn{2}{c}{0.382}          & \multicolumn{2}{c|}{\textbf{0.413}} & \multicolumn{2}{c|}{0.583}                & \multicolumn{2}{c}{{\ul \textbf{0.625}}}
\end{tabular}
}
			    \vspace{-0.3cm}
    \caption{\textbf{Comparisons on HPatches~\cite{balntas2017hpatches}.} HEstimation, Precision and Recall are calculated at the threshold of 3 pixels. The best score among methods is underlined and the better one between losses is in bold. }
	\label{tbl:hpatches}
		    \vspace{-0.4cm}
\end{table}
}
\vspace{-0.8cm}
\subsubsection{Image Matching} \label{sec:2D Feature Matching}
\vspace{-0.1cm}
In the image matching experiment, we use the HPatches dataset ~\cite{balntas2017hpatches}, which has been widely adopted to evaluate the quality of image matching ~\cite{mishchuk2017working, detone2018superpoint, revaud2019r2d2, liu2019gift, tian2019sosnet, pautrat2020online, wiles2020d2d}. Following D2-Net ~\cite{dusmanu2019d2} and ASLFeat \cite{luo2020aslfeat}, we exclude $8$ high-resolution sequences, leaving $52$ and $56$ sequences with illumination or viewpoint variations, respectively. For a precise reproduction, we directly use the open source code of two state-of-the-art joint description and detection of local features methods, ASLFeat and D2-Net, replacing their losses with ours. \wb{SuperPoint (SP)~\cite{detone2018superpoint} is also a powerful approach to imagine matching. However, it resorts to interest point pre-training and self-labelling that need synthetic shapes and homographic adaptation, which are very difficult to be directly adopted with our loss. Despite this, we still report the 2D matching results by SuperPoint in Tab.~\ref{tbl:hpatches} to better present the enhancements on other baselines.} Particularly, we keep the same evaluation settings as the original papers for both training and testing.
\PAR{Results on the HPatches.} Here, three metrics~\cite{pautrat2020online} are used: 1) Homography estimation (HEstimation), the percentage of correct homography estimation between an image pair; 2) Precision, the ratio of correct matches over possible matches; 3) Recall, the percentage of correct predicted matches over all ground truth matches. As illustrated in Tab. ~\ref{tbl:hpatches}, when using our loss, clear improvements (up to $3.9\%$) under illumination variations can be seen in almost all metrics. The only exception happens for D2-Net on Recall and ASLFeat on HEstimation where our loss is only negligibly inferior. On the other side, the performance gain from our method can be observed on all metrics under view variations. This gain ranges from $1.2 \%$ to $5.6 \%$. Our proposed optimization strategy shows more significant improvements under view changes than illumination changes.

\vspace{-0.3cm}
\subsubsection{Point Cloud Registration}
\vspace{-0.1cm}
\begin{table}[t]
\Huge
    \centering
    \resizebox{\linewidth}{!}{
\begin{tabular}{l||ccc|ccc|ccc}
          & \multicolumn{3}{c|}{FCGF \cite{choy2019fully}}                                                    & \multicolumn{3}{c|}{D3\_Contrastive \cite{bai2020d3feat}} & \multicolumn{3}{c}{\textbf{D3\_Ours}} \\
          & Reg  & FMR                               & IR                            & Reg        & FMR       & IR      & Reg     & FMR    & Inlier   \\ \hline \hline
Kitchen   & 0.93 & \multirow{8}{*}{\textbackslash{}} & \multirow{8}{*}{\textbackslash{}} & 0.97       & 0.97      & 0.34        & \textbf{0.98} & \textbf{0.99} & \textbf{0.46} \\
Home 1    & 0.91 &                                   &                                   & 0.90       & 0.99      & 0.45        & \textbf{0.92} & \textbf{1.00}  & \textbf{0.59} \\
Home 2    & 0.71 &                                   &                                   & 0.72       & 0.91      & 0.43        & \textbf{0.73} & \textbf{0.93} & \textbf{0.55} \\
Hotel 1   & 0.91 &                                   &                                   & 0.95       & 0.98      & 0.39        & \textbf{0.98} & \textbf{1.00}  & \textbf{0.53} \\
Hotel 2   & 0.87 &                                   &                                   & 0.87       & 0.95      & 0.37        & \textbf{0.91} & \textbf{0.97} & \textbf{0.49} \\
Hotel 3   & 0.69 &                                   &                                   & 0.80       & 0.96      & 0.47        & \textbf{0.81} & \textbf{1.00}  & \textbf{0.56} \\
StudyRoom & 0.75 &                                   &                                   & 0.83       & 0.95      & 0.37        & \textbf{0.86} & \textbf{0.96} & \textbf{0.56} \\
MIT Lab   & 0.80 &                                   &                                   & 0.69       & 0.92      & 0.42        & \textbf{0.84} & \textbf{0.97} & \textbf{0.54} \\ \hline
Average   & 0.82 & 0.95                              & \textbf{0.54}                     & 0.84       & 0.95      & 0.41        & \textbf{0.88} & \textbf{0.98} & \textbf{0.54}  
\end{tabular}
}
			    \vspace{-0.25cm}

    \caption{\textbf{Comparisons on 3DMatch~\cite{zeng20173dmatch}}. Reg, FMR and IR are evaluated at the threshold of $0.2$ m, $5 \%$ and $0.1$ m.}
	\label{tbl:3dmatch}
\vspace{-0.3cm}
\end{table}

In terms of 3D domain, we use the 3DMatch~\cite{zeng20173dmatch}, a popular indoor dataset for point cloud matching and registration ~\cite{khoury2017learning, deng2018ppfnet, gojcic2019perfect, choy2019fully, choy2020high, gojcic2020learning, choy2020deep}. We follow the same evaluation protocols in \cite{zeng20173dmatch} to prepare the training and testing data, 54 scenes for training and the remaining 8 scenes for testing. As D3Feat\cite{bai2020d3feat} is the only work which jointly detects and describes 3D local features, we replace its loss with ours for comparison. To better demonstrate the improvements, the results from FCGF \cite{choy2019fully} are also included.
\vspace{-0.1cm}
\PAR{Results on the 3DMatch.} We report the performance on three evaluation metrics: 1) Registration Recall (Reg), 2) Inlier Ratio (IR), and 3) Feature Matching Recall (FMR). As illustrated in Tab.~\ref{tbl:3dmatch},  when our P2-Loss is adopted, a $4 \%$ and a $3 \%$ improvements can be seen on Reg and FMR, respectively. In contrast, there is only $2 \%$ and $0 \%$ respective difference between FCGF and the original D3Feat. In particular, as for Inlier Ratio, our loss demonstrates better robustness, outperforming the original one by $13 \%$, comparable to FCGF. Overall, P2-Loss consistently achieves the best performance in terms of all metrics.

\subsection{The Impact of Descriptor Loss}
\vspace{-0.1cm}
\label{sec:discussion} 
Finally, we come to analyse the impacts of loss choices on homogeneous (2D$\leftrightarrow$2D or 3D$\leftrightarrow$3D) and heterogeneous (2D$\leftrightarrow$3D) feature matching. From the detector loss formulation in Eq.~\ref{eq:detector}, we can see that its optimization tightly depends on the descriptor. Therefore, we conduct a comprehensive study on three predominant metric learning losses for descriptor optimization and aim to answer: why is the circle-guided descriptor loss best suited for feature matching? To this end, we track the difference between the positive similarity $d_p$ and the most negative similarity $d_{n^*}$ ($\max$($d_{n}$)) with various loss formulations and architectures. 

\begin{figure}
    \centering
    \includegraphics[width=0.93\columnwidth]{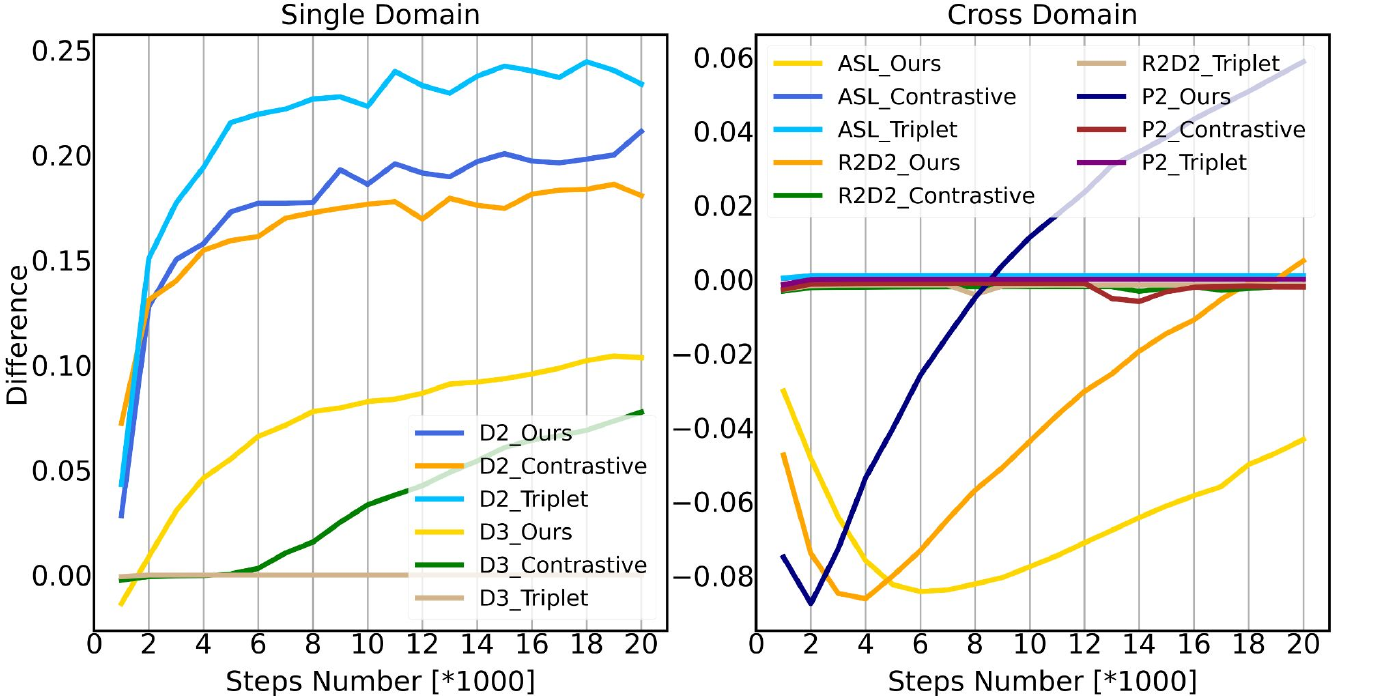}
        		    \vspace{-0.18cm}
    \caption{The difference between the positive similarity $d_p$ and the most negative similarity $d_{n^*}$ over time with different networks and losses. Left: single-domain matching; Right: cross-domain matching.}
    \label{fig:similarity}
    		    \vspace{-0.2cm}
\end{figure}

Fig.~\ref{fig:similarity} (left) shows that, in single/homogeneous 2D or 3D domains, both D2-Net and D3Feat can gradually learn distinctive descriptors. D2-Net consistently ensures convergence, regardless of the choice of loss, while D3Feat fails when hard-triplet loss is selected. This is consistent with the conclusion in \cite{bai2020d3feat}. In the cross-domain image and point cloud matching (Fig.~\ref{fig:similarity} (right), we compare different losses and 2D feature extractors. This overwhelmingly demonstrates that neither hard-triplet nor hard-contrastive loss can converge in any framework (ASL, R2D2 or P2-Net).
Both triplet and contrastive losses are inflexible, because the penalty strength for each similarity is restricted to be equal. Moreover, their decision boundaries are parallel to $d_p$=$d_n$, which causes ambiguous convergence \cite{chopra2005learning, mishchuk2017working}. However, our loss enables all architectures to converge, showing promising trends towards learning distinctive descriptors. Thanks to the introduction of \emph{circular decision boundary}, the proposed descriptor loss assigns different gradients to the similarities, promoting more robust convergence \cite{sun2020circle}.
% During training, the gradient back-propagated to $d_p^i$ ($d_n^j$) will be amplified by $[O_p-d_p^i]_+$ ($[d_n^j-O_n]_+$). Those less-optimized similarity scores will have larger weighting factors and consequentially get larger gradients.

Interestingly, we can observe that the distinctiveness of descriptors initially is inverted for heterogeneous matching, unlike homogeneous matching. As pixel and point descriptors are initially disparate, their similarity can be extremely low for both positive and negative matches in the initial phase\footnote{Please refer to the supplementary material for more analysis.}. In such case, the gradients (ranging between $[0, 1]$) with respect to $d_p$ and $d_n$ almost approach 1 and 0 \cite{sun2020circle}, respectively. Because of the sharp gradient difference, the loss minimization in network training will tend to over-emphasize the optimization $d_p$ while sacrificing the descriptor distinctiveness. As $d_p$ increases, our loss reduces its gradient and thus enforces a gradually strengthened penalty on $d_n$, encouraging the distinctiveness between $d_p$ and $d_n$.
%  unlike the case in single domain, 

\vspace{-0.1cm}
\section{Conclusions}
\vspace{-0.1cm}
In this work, we propose P2-Net, a dual and fully-convolutional framework in combination with an ultra-wide reception mechanism to jointly describe and detect 2D and 3D local features for direct matching between pixels and points. Moreover, a novel loss function, P2-Loss that consists of a circle-guided descriptor loss and a batch-hard detector loss, is designed to explicitly guide the network to learn distinctive descriptors and detect repeatable keypoints for both pixels and points. Extensive experiments on pixel and point matching, visual localization, image matching and point cloud registration not only show the effectiveness and practicability of our P2-Net but also demonstrate the generalization ability and superiority of our P2-Loss.

\clearpage
{\small
\bibliographystyle{ieee_fullname}
\bibliography{main}
}

\end{document}